\documentclass[10pt,journal,compsoc]{IEEEtran}
\usepackage{float}
\usepackage{mathtools}
\usepackage{framed}
\usepackage{multirow}
\usepackage[table,xcdraw]{xcolor}
\usepackage[draft]{pgf}
\usepackage{duckuments}
\usepackage{graphicx}
\usepackage{url}
\usepackage{xcolor}
\usepackage[english]{babel}
\usepackage{cite}
\usepackage{amssymb}
\usepackage{booktabs}
\usepackage{hyperref}
\usepackage{blindtext}
\usepackage{stfloats}
\usepackage{lipsum}

\ifCLASSINFOpdf
\else
\fi
\begin{document}
% \title{Detecting natural disasters, damage, and incidents in the wild}
\title{Incidents1M: a large-scale dataset of images with natural disasters, damage, and incidents}

\author{Ethan~Weber,
        Dim~P.~Papadopoulos,
        Agata~Lapedriza,
        Ferda~Ofli,
        Muhammad~Imran,
        Antonio~Torralba
\IEEEcompsocitemizethanks{\IEEEcompsocthanksitem E. Weber was with CSAIL at MIT, Cambridge, MA. He is now a PhD student at BAIR in Berkeley, CA.\protect\\\vspace{-9px}
\IEEEcompsocthanksitem D. Papadopoulos was with CSAIL at MIT, Cambridge, MA. He is now an Assistant Professor at the Technical University of Denmark (DTU).
\IEEEcompsocthanksitem A. Lapedriza is a professor at Universitat Oberta de Catalunya (Barcelona, Spain) and a Research Affiliate at MIT Medialab, Cambridge, MA.\protect\\\vspace{-9px}
\IEEEcompsocthanksitem A. Torralba is with CSAIL.\protect\\\vspace{-9px}
\IEEEcompsocthanksitem F. Ofli and M. Imran are with QCRI in Doha, Qatar.\protect\\\vspace{-9px}
\IEEEcompsocthanksitem E-mails: %
\{ejweber@mit.edu,ethanweber@berkeley.edu\}, \{dimpapa@mit.edu,dimp@dtu.dk\}, \{alapedriza@uoc.edu,agata@mit.edu\},  torralba@mit.edu, \{fofli, mimran\}@hbku.edu.qa}
}

\newcommand{\todo}[1]{\textcolor{red}{[#1]}}

\iftrue     
%\iffalse
\newcommand{\dimpp}[1]{\textcolor{blue}{[DP: #1]}}
\newcommand{\antonio}[1]{\textcolor{magenta}{[AT: #1]}}
\newcommand{\ethan}[1]{\textcolor{cyan}{[EW: #1]}}
\newcommand{\ferda}[1]{\textcolor{green}{[FO: #1]}}
\else
\newcommand{\dimpp}[1]{\textcolor{blue}{\noindent}}
\newcommand{\antonio}[1]{\textcolor{magenta}{\noindent}}
\newcommand{\ethan}[1]{\textcolor{cyan}{\noindent}}
\newcommand{\ferda}[1]{\textcolor{green}{\noindent}}
\fi

\newcommand{\mypar}[1]{\vspace{1mm}\noindent{\textbf{#1}}}

% https://tex.stackexchange.com/questions/9363/how-does-one-insert-a-backslash-or-a-tilde-into-latex
\newcommand{\textapprox}{\raisebox{0.5ex}{\texttildelow}}

\newcommand\blfootnote[1]{%
  \begingroup
  \renewcommand\thefootnote{}\footnote{#1}%
  \addtocounter{footnote}{-1}%
  \endgroup
}

\IEEEtitleabstractindextext{%

\begin{abstract}
Natural disasters, such as floods, tornadoes, or wildfires, are increasingly pervasive as the Earth undergoes global warming. It is difficult to predict when and where an incident will occur, so timely emergency response is critical to saving the lives of those endangered by destructive events. Fortunately, technology can play a role in these situations. Social media posts can be used as a low-latency data source to understand the progression and aftermath of a disaster, yet parsing this data is tedious without automated methods. Prior work has mostly focused on text-based filtering, yet image and video-based filtering remains largely unexplored. In this work, we present the Incidents1M Dataset, a large-scale multi-label dataset which contains 977,088 images, with 43 incident and 49 place categories. We provide details of the dataset construction, statistics and potential biases; introduce and train a model for incident detection; and perform image-filtering experiments on millions of images on Flickr and Twitter. We also present some applications on incident analysis to encourage and enable future work in computer vision for humanitarian aid. Code, data, and models are available at \url{http://incidentsdataset.csail.mit.edu}.
\end{abstract}


}

\maketitle

\IEEEdisplaynontitleabstractindextext

% \fi
\IEEEpeerreviewmaketitle

% ########################################################
\IEEEraisesectionheading{\section{Introduction}\label{sec:introduction}}

\IEEEPARstart{E}{fficient} detection of natural disasters, such as floods, wildfires, and other events that require human intervention, is essential for relief organizations. Particularly, in the occurrence of an emergency, rapidly acquiring information is a key factor to organize an optimized response. Unfortunately, in the case of events that need human intervention, information analysis still requires a lot of manual processing, which is costly and often inefficient. Recently, there have been some efforts on using computer vision techniques on satellite imagery, synthetic aperture radar, and other remote sensing data~\cite{gamba_rapid_2007,plank2014rapid,skakun2014flood,chehata2014object} to overcome the need of manual data processing. However, these efforts do not allow to fully automate the data processing yet, since they are not robust enough. Furthermore, satellite imagery only provides an overhead view of the area and is often affected by occlusions caused by clouds (which are common in the case of critical weather conditions) and smoke (which is common in natural disaster including wildfires, volcanic eruptions, or hurricanes).

In this work, we explore how to automate information processing about natural disasters from a different source: social media posts. Studies show that right after a disaster occurs, social media contains relevant information to disaster response, such as reports of damages or urgent needs of affected people, among others~\cite{castillo2016big,imran2015processing}. In particular, this information is often in the form of images and videos. Unlike data sources such as satellite imagery, exploiting the information in on-the-ground social media images for relief organization response had remained unexplored until very recently in our prior work~\cite{weber2020detecting}, mainly because of technical challenges. Automatically filtering relevant images is challenging, since image streams on social media are very noisy and a large percentage of images posted are not related to humanitarian needs.  Also, automatic filtering of relevant images requires robust deep learning models, which need to be trained with large amounts of labelled images. However, creating a suitable large-scale labeled dataset for the task of incident recognition in the wild is costly.

\begin{figure*}[t]
\centering
\includegraphics[width=\linewidth]{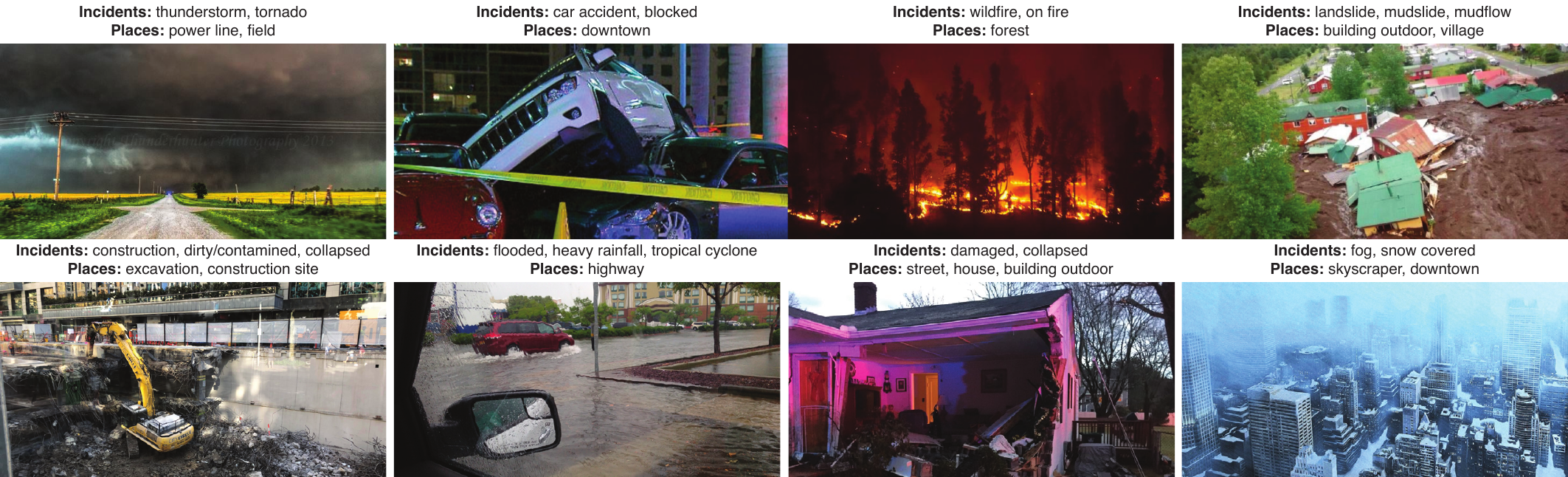}
\caption{\textbf{Incidents1M Dataset images.} These are some images from the Incidents1M Dataset, with their corresponding incident and place labels. 275,381 images have more than one incident labels. The place labels provide additional context for where the incident is occurring.}
\label{fig:incidents_dataset_examples}
\end{figure*}

This paper presents the large-scale, multi-label Incidents1M Dataset (Fig.~\ref{fig:incidents_dataset_examples}). We extend the Incidents Dataset~\cite{weber2020detecting} from 446,684 to 977,088 images and remove the single-label assumption (of having at most one incident and place category per image). Instead we obtain multiple labels per image. In this paper we explain the process for creating this dataset, analyze its statistics and potential biases, train a new Incidents1M Model and demonstrate applications of detecting incidents in the wild. We then explore additional applications in damage assessment by training an off-the-shelf generative model~\cite{Karras2019stylegan2} and discuss how a real-time incident monitoring dashboard may be useful. We expect that the larger Incidents1M Dataset with multi-labels will enable more insights and applications beyond its original introduction~\cite{weber2020detecting}.

% ########################################################
% ########################################################
% ########################################################
% ########################################################
\section{Related Work}
\label{sec:rel_work}

\mypar{Computer vision for social good.}
Many vision-based technologies fall short of reaching out to diverse geographies and communities due to biases in the commonly-used datasets. For instance, state-of-the-art object recognition models perform poorly on images of household items found in low-income countries~\cite{vries_does_2019}. To remedy this shortcoming, the community has made recent progress in areas including mapping economic development and poverty~\cite{jean_combining_2016,piaggesi_predicting_2019,tingzon_mapping_2019,helber_mapping_2019,watmough_socioecologically_2019}, 
creating high-resolution maps in the developing world~\cite{nachmany_detecting_2019,bonafilia_building_2019}, 
recognizing displaced people and human rights violations~\cite{kalliatakis_displacenet_2019,kalliatakis_exploring_2019}, 
assessing demographic makeup and perceived safety of urban areas~\cite{naik_streetscore_2014,gebru_using_2017}, 
estimating population density and urban analytics~\cite{arietta2014city,gebru2017fine,workman2017unified,can2018ambiance,kataoka2019ten,naik_streetchange_2017,zhou2014recognizing}, 
improving health decisions~\cite{wu_deep_2019,rehman_deep_2019,mckinney_international_2020}, 
crop monitoring and food security~\cite{pryzant_monitoring_2017,efremova_aibased_2019,rustowicz_semantic_2019,kaneko_deep_2019}, 
climate change~\cite{schmidt_visualizing_2019}, and wildlife preservation~\cite{kellenberger_when_2019}. 
These studies and others have shown the potential of computer vision to create impact for social good at a global scale.

\mypar{Incident detection in satellite imagery.}
There are numerous studies that combine traditional machine learning with a limited amount of airborne or satellite imagery over disaster zones
~\cite{turker_detection_2004,gamba_rapid_2007,chehata2014object,skakun2014flood,radhika2015cyclone,FernandezGalarreta:2015dn}. For a detailed survey, see~\cite{joyce2009review,dell_remote_2012,dong_comprehensive_2013,plank2014rapid}. Oftentimes, these studies are constrained to particular disaster events. However, recently deep learning-based techniques have been applied on larger collections of airborne or remote-sensed data to  
detect damaged buildings~\cite{gueguen2015large,attari2016nazr,li2019building,xu_building_2019,gupta_creating_2019,weber2020building}, 
segment flooded regions~\cite{nogueira2018exploiting,ben-haim_inundation_2019,rudner_multi3net_2019},
estimate extent of fires~\cite{radke2019firecast,doshi_firenet_2019},
assess hurricane destruction~\cite{li2018semisupervised,sublime2019automatic}, 
perform fine-grained analysis of disaster scenes~\cite{chowdhury2020comprehensive,rahnemoonfar2020floodnet}, 
and compute a disaster impact index~\cite{doshi_from_2018}.
Other studies have applied transfer learning~\cite{seo_revisiting_2019} and few-shot learning~\cite{oh_explainable_2019} to deal with unseen situations during disasters.

\mypar{Incident detection in social media.}
More recently, social media has emerged as a non-traditional data source for rapid disaster response. Most studies have focused heavily on text messages for extracting crisis-related information~\cite{imran2015processing,reuter2018fifteen}. On the contrary, there are only a few studies using social media images for disaster response~\cite{peters2015investigating,chen2013understanding,daly2016mining,nguyen2017damage,nia2017building,alam2019processing,li2019identifying,pouyanfar2019unconstrained,Abavisani_2020_CVPR,alam2020deep,chaudhary2020water,weber2020detecting}. For example, \cite{nguyen2017damage} classifies images into three damage categories whereas \cite{nia2017building} regresses continuous values indicating the level of destruction. Recently, \cite{alam2019processing} presented a system with duplicate removal, relevancy filtering, and damage assessment for analyzing social media images. \cite{chaudhary2020water,quan2020flood} developed a method to predict flood water level from social media images. \cite{li2019identifying,pouyanfar2019unconstrained} investigated domain adversarial networks to cope with data scarcity during an emergent disaster event.

\mypar{Incident detection datasets.}
Most of the aforementioned studies use small datasets covering just a few disaster categories, which limits the possibility of creating methods for automatic incident detection. In addition, the reported results are usually not comparable due to lack of public benchmark datasets, whether it be from social media or satellites~\cite{disaster_detection_survey}. One exception is the xBD dataset~\cite{gupta2019xbd}, which contains 23,000 images annotated for building damage but covers only six disasters types (earthquake, tsunami, flood, volcanic eruption, wildfire, and wind). On the other hand, \cite{gueguen2015large} has many more images but their dataset is constructed for detecting damage using pre- and post-disaster images. There are also datasets combining social media and satellite imagery for understanding flood scenes~\cite{mediaeval_2017,mediaeval_2018} but they have up to 11,000 images only. In summary, existing incident datasets are small, both in number of images and categories. In particular, incident datasets are far, in size, from the current large datasets on image classification, like ImageNet~\cite{imagenet_2009} or Places~\cite{places_dataset}, which contain millions of labeled images. Unfortunately, neither ImageNet nor Places covers incident categories. Extending our previous work~\cite{weber2020detecting}, our dataset is significantly larger and more diverse than any other available dataset related to incident detection.

\mypar{Out-of-distribution detection.}
Classification and object detection models trained and tested on the same data distribution achieve remarkable results. However, these models often suffer when deployed on out-of-distribution data. For classification~\cite{imagenet_2009,places_dataset}, the task is to select the correct class with the implicit assumption that at least one of the classes will be present in the image. For object detection~\cite{Girshick_2014_CVPR,Lin_2017_ICCV,Shrivastava_2016_CVPR}, the task is to first filter for whether or not an object exists, and then this is typically followed by classification. Our setting is closer to object detection rather than classification because we aim to filter many social media images to identify the very few relevant images with incidents. To avoid false-positive predictions with classification and detection models, prior work has focused on improving training to incorporate negatives~\cite{Durand_2019_CVPR,lee2018training} or adjusting model confidence at test time~\cite{liang2017enhancing,leeNIPS2018}. Keeping this in mind, we choose to include and collect class-negative labels (or, hard-negatives) for our dataset to enable training a robust model for incident detection.

\begin{figure*}[t]
\centering
\includegraphics[width=\linewidth]{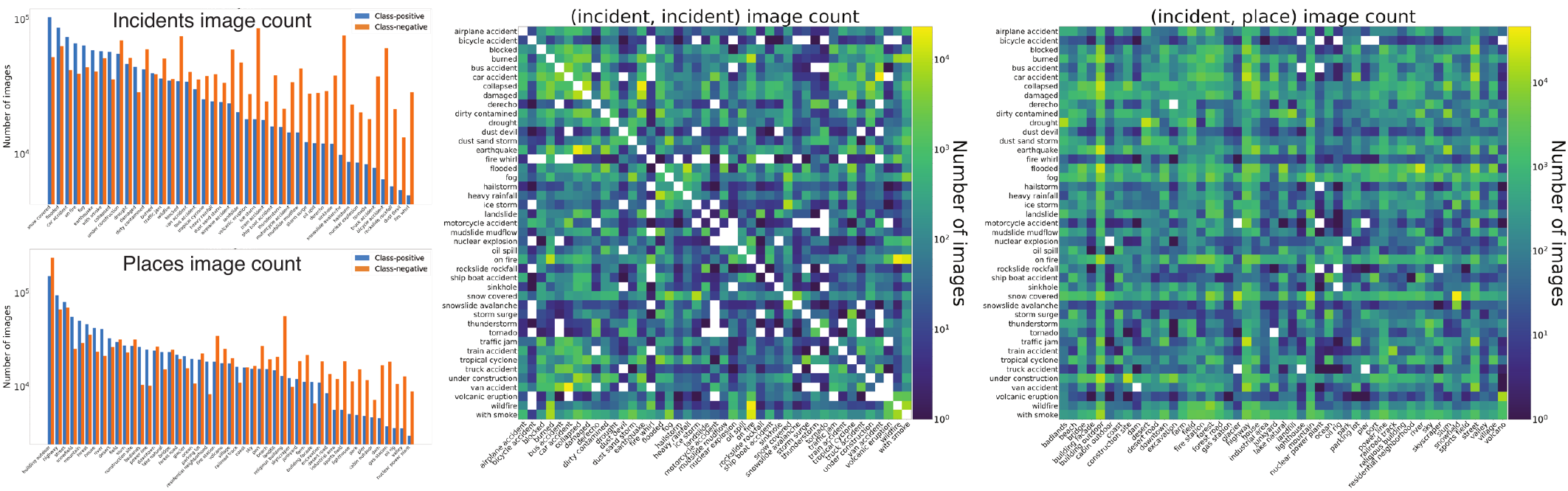}
\caption{\textbf{Dataset statistics}. (Left) The number of class-positive and class-negative images for the incident and place categories. (Middle) The number of images that contain both class-positive incidents, excluding the same-incident diagonal. This highlights interesting incident correlations (e.g., ``earthquake" and ``collapsed"). (Right) The number of images with both a class-positive incident and place label (e.g., ``traffic jam" and ``highway").}
\label{fig:dataset_statistics}
\end{figure*}

% ########################################################
\section{Incidents1M Dataset}
\label{sec:dataset}

In this section, we present the Incidents1M Dataset, a multi-label dataset containing many images of incident-related scenes across a variety of locations. Here we define the incident and place categories, explain the image downloading and labeling process, and present the final dataset statistics.

\subsection{Incident and place categories} The Incidents1M Dataset contains both incident and place categories. For the incident categories, we first consider an extensive list found online\footnote{\url{https://en.wikipedia.org/wiki/Disaster}} and create a fine-grained list of 233 incident categories. We manually merge visually-similar categories (e.g., ``snow storm" and ``blizzard", and ``mudslide" and ``mudflow") and remove categories unlikely to be recognized (e.g., ``heat wave", ``infestation", ``famine"). This results in 43 incident categories. Similarly, for places we first consider the 118 outdoor categories of Places Dataset~\cite{places_dataset} and merge categories in the same super-category. This results in 49 place categories. All category names are shown in Fig.~\ref{fig:dataset_statistics}.

\subsection{Image downloading and duplicate removal}

With the 43 incidents and 49 places defined, we use the pairwise combinations to create 43 x 49 = 2,107 (incident, place) pairs. Using synonyms and prepositional phrases, we amplify and convert these pairs into a total of 10,188 Google Image queries (e.g., ``car accident in highway'' and ``car wreck in flyover'', or ``blizzard in street'' and ``snow storm in alley''). We then download all images returned from Google for each query. This results in 6,178,192 images. To remove duplicate images, we extract deep features with a pre-trained model (ResNet18~\cite{he2016deep} on Places~\cite{zhou2017scene}) and cluster images with a nearest neighbors algorithm. We remove extra images and keep 3,487,339 images to consider for labeling.

\subsection{Multi-label category annotation}
\label{sec:annotation}

In previous work \cite{weber2020detecting}, we labeled these images with up to one incident and one place per image. However, here we remove the single-label assumption and instead label images with multiple categories where applicable. Multiple labels per image is more accurate, e.g., a building may be both ``on fire" and ``collapsed"; our incident and place categories are not mutually exclusive.

\mypar{Labeling tasks.} The downloaded images from Google are noisy, so we use the Amazon Mechanical Turk (MTurk) platform to manually obtain class labels. Similar to \cite{places_dataset}, we create Human Intelligence Tasks (HITs) where human annotators are shown 100 images sequentially and asked binary questions of whether the images have a particular incident or place label (e.g., Does this image contain a ``wildfire"? or, Does this image occur in a ``forest"?). A total of 15 images (10 positive and 5 negative) out of the 100 in a HIT are used as quality control. We only keep labels for HITs with annotator accuracy above 85\% on the 15 control images.

\begin{figure}[t]
\centering
\includegraphics[width=\linewidth]{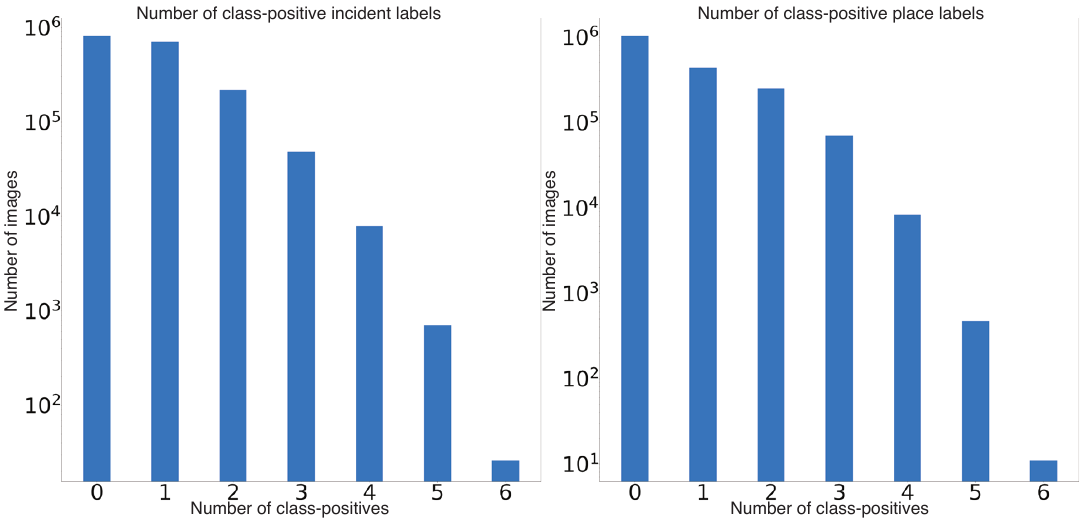}
\caption{\textbf{Multi-label statistics.} The Incidents1M Dataset has multiple labels per image, both class-positive and class-negative. Some images have up to 6 class-positive incident labels. (Left) We show the number of images that have the specified number of class-positive labels. (Right) We show the same information but for place labels. (Left and right) Notice that the 0 columns indicate the number of class-negative images, where no labeled incidents or places occur.}
\label{fig:multi_label_frequency}
\end{figure}

\mypar{Class-negative labels.} Whenever a human annotator is asked a binary question, we obtain either a class-positive (YES response) or a class-negative (NO response) label. Both annotations are included in our dataset because they provide information about whether or not the particular incident or place appears in the image. Furthermore, class-negative labels are particularly useful for training a detection model that can perform in the wild. Concretely, we use class-negatives to mitigate false positive predictions (see Sec.~\ref{sec:training_with_class_negatives} for more details).

\mypar{Labeling procedure.} To start the multi-labeling procedure, we run incident HITs for \textapprox1M of the initial 3,487,339 images. We ask the binary questions for the category used in the original Google query, and this results in a set of class-positive and class-negative incident images. For the class-positive incident images (those with at least one incident), we then run HITs to obtain place labels in the same way. Next, we train a model (Sec.~\ref{sec:incidents_model}) on the labeled images, and we run inference on the remaining images to assign a confidence score to every image for all 43 incident and 49 place categories. We sort these confidence scores in decreasing order, and we create HITs for the top ranked incident categories to obtain labels. For any image that has at least one class-positive incident label, we create and run HITs for the top ranked place categories. In prior work~\cite{weber2020detecting}, we stopped labeling when an image had at least one class-positive incident. Here, we continue the process and allow for annotators to label an image more than once, up to 6 times. This makes the Incident1M Dataset significantly larger and more useful.

\subsection{Dataset statistics}

The Incidents1M Dataset contains 1,787,154 images, where 977,088 are class-positive incident images (having at least one class-positive label) and 810,066 are class-negative incident images (having no class-positive labels but at least one class-negative label). For places, 764,124 images are class-positive and 1,023,030 are class-negative. We show a more complete breakdown of these numbers by categories, as well as the frequency of incidents co-occurring with other incidents or places in Fig.~\ref{fig:dataset_statistics}. Furthermore, the class-positive incident images have a mean of 1.35 class-positive labels per image. For class-positive place images the mean is 1.55. We report additional statistics on our multi-label statistics in Fig.~\ref{fig:multi_label_frequency}, where we show the number of images with a specific number of class-positive labels, for both incidents (left) and places (right).

% ########################################################
\section{Incidents model}
\label{sec:incidents_model}
In this section, we introduce the Incidents1M Model for detecting incidents in images. We train the model with all 1,787,154 images, which have at least one class-positive or class-negative label.

\subsection{Multi-class multi-label architecture}
We consider a multi-class multi-label setting because we have 43 incident and 49 place categories, and multiple labels can co-occur together (Fig. \ref{fig:dataset_statistics}). Furthermore, incidents and places can be visually related and share contextual information, so we follow the multi-task learning paradigm \cite{caruana1997multitask,rebuffi2017learning,simonyan2014two} and employ a Convolutional Neural Network (CNN) with a shared backbone and two task-specific output branches. The incident and place branches pass through a sigmoid layer and are transformed to tensors of the form $\mathbb{R}^{43}$ and $\mathbb{R}^{49}$, respectively. Each value in the vectors has a confidence score in the range $(0,1)$.

\subsection{Training with a class-negative loss}
\label{sec:training_with_class_negatives}
Our setting is more similar to incident detection than to incident classification. With our interest in in-the-wild data such as on Twitter, the number of images depicting an incident is very little compared to the vast majority of images. Much like the well-known problem of object detection~\cite{Girshick_2014_CVPR,Lin_2017_ICCV,Shrivastava_2016_CVPR}, we want the Incidents1M Model to detect positive incident examples from a pool of many images, including challenging negative images. For instance, a chimney with smoke or a fireplace are not disaster situations, yet they share visual features similar to our ``with smoke'' and ``on fire'' incident categories. Prior work exists to mitigate false-positive detections by either modifying the training process~\cite{Durand_2019_CVPR,lee2018training} or adjusting model confidence scores during test time~\cite{liang2017enhancing,leeNIPS2018}. We choose to do the former and modify our training process to incorporate class-negative images, or images where we have information that no incident of a particular category exists.

We formulate the loss by modifying a binary cross entropy (BCE) loss similar to \cite{Durand_2019_CVPR} to use partial labels from the 1,787,154 images. More specifically, we incorporate class-negative information into the loss to mitigate false positive predictions. The class-negative images are, in-fact, hard-negatives given the way that we annotated images described in Sec.~\ref{sec:annotation} (i.e., asking questions based on Google queries or prioritizing questions based on high confidence scores returned from our model). For each task-specific output branch, our loss takes the same form. The incident loss $\mathcal{L}_{in}$ is formalized as follows:

\begin{equation}\label{eq:loss}
\mathcal{L}_{in} = \sum_{i=1}^N w_i \Big( y_i \log(x_i) + (1 - y_i) \log(1 - x_i)  \Big)
\end{equation}

\noindent where $N = 43$ is the number of classes, $x$ is the predicted class probabilities, $y$ is the target class labels, and $w$ is the weight vector indicating where we have information ($w_i = 1$ when we have either a class-positive or class-negative label, otherwise $w_i = 0$). As a way to bias the model to suppress false confidence scores, we set all $\{{w_i}\}_{i=1}^N = 1$ when $\sum y_i > 0$, meaning at least one class-positive label exists. The place loss $\mathcal{L}_{pl}$ takes the same form with $N=49$, and the total loss $\mathcal{L} = \mathcal{L}_{in} + \mathcal{L}_{pl}$.

% ########################################################
\section{Experiments within the Incidents Dataset}
\label{sec:experiments}

In this section, we run experiments using our multi-label Incidents1M Dataset. We explore the importance of the class-negative loss and also the effect of the dataset size. We use our best model for the remainder of the paper.

\subsection{Dataset splits and evaluation setting}

In this work, we introduce the Incidents1M Dataset, which is an extension of the Incidents Dataset~\cite{weber2020detecting}. The Incidents1M Dataset contains \textapprox2x more images and has multiple class-positive labels per image. In this section, we train and evaluate our models with ``Incidents1M Dataset''. We split the dataset into a train (90\%), val (5\%), and test (5\%) set. Each split of the ``Incidents1M Dataset'' is a larger and more complete superset of the corresponding split from the original ``Incidents Dataset''~\cite{weber2020detecting}. We evaluate the mAP for both the incident and place categories separately.

\mypar{Implementation details.} We use a ResNet-50~\cite{he2016deep} for the model backbone, which has been pretrained on the Places dataset~\cite{places_dataset} used for scene classification. The incident and place branches are each a fully-connected layer mapping from 2048 dimensions to the 43 and 49 dimensions for the incident and place categories, respectively. For every model, we train until convergence with early stopping based on the average incidents and places mAP on the hold-out validation set. We use an image batch size of 256, the Adam optimizer with a learning rate of 1e-4, and at most 20 epochs. Note that because our dataset is substantially larger than our first introduction, we no longer use additional images from the Places~\cite{places_dataset} dataset as was done in~\cite{weber2020detecting}.

\begin{figure*}[t]
\centering
\includegraphics[width=\linewidth]{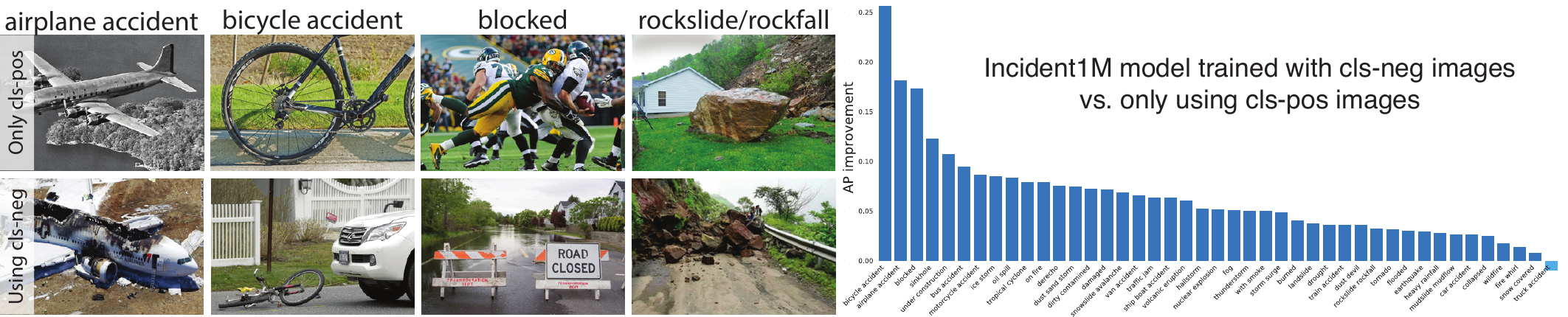}
\caption{\textbf{Class-negative usefulness.} (Top left) We show images scored highly for the corresponding incident when trained with only class-positive labels. (Bottom left) We show highly ranked images when both class-positive and class-negative labels are used. Notice that the model using class-negatives is more robust to similarly looking visual features or noisy labels (e.g., incorrect labeling of ``blocked" due to its polysemous usage in sports context). (Right) We show the per-class AP improvement for both models, with and without class-negatives. AP increases for almost all incidents.}
\label{fig:class_negatives}
\end{figure*}

\subsection{Class-negative usefulness}
\label{sec:class-negatives}
First we show that using class-negatives improves model performance. In particular, we train a model with and without using class-negative images. When using class-positive (cls-pos) images only, this means that we use the 977,088 images with at least one class-positive label and do not incorporate any class-negative (cls-neg) labels into the loss described in Eq.~\ref{eq:loss}. When we use the class-negatives, the loss is used normally and high confidence scores are penalized where we have class-negative information. The top two rows of Tab.~\ref{table:class_negatives_and_larger_dataset} illustrate that the incident mAP goes from $62.83$ to $67.19$, i.e., a \textapprox7\% increase in performance. The place mAP goes down only by a negligible amount; this may be attributed to class-negatives for incidents being more applicable than class-negatives for places. For instance, ``not a car accident'' describing an image with an intact car is very informative to a model learning visual features. Similar examples for place labels are harder to construct.

In Fig.~\ref{fig:class_negatives}, we show (left) qualitative results and (right) per-class incident AP improvements when using the class-positive only model (cls-pos) vs. using both (cls-pos \& cls-neg). The images shown are some of the highest scored images from each model. Notice that when class-negatives are not used, the model has never seen an intact airplane or an intact bicycle. Furthermore, the model may pick up on any amount of noise that exists from incorrect labels in the dataset. Even though we take extensive measure to ensure high quality in the dataset, some incorrect labels may appear (e.g., the polysemous term ``blocked" referring to American football rather than an incident, shown in column 3 of the figure). When the model is trained with class-negatives, however, it becomes more robust to similar features (e.g., the intact wings of the airplane) or incorrect labels, and reliably scores true incidents with higher confidence while mitigating false positive predictions. We see that all incident categories besides ``truck accident" experience an improvement in AP; we attribute the decrease in the ``truck accident" score to both the small number of ``truck accident" labels and possible confusion with other categories, namely ``bus accident", ``car accident", or ``van accident".

\begin{figure*}[t]
\centering
\includegraphics[width=\linewidth]{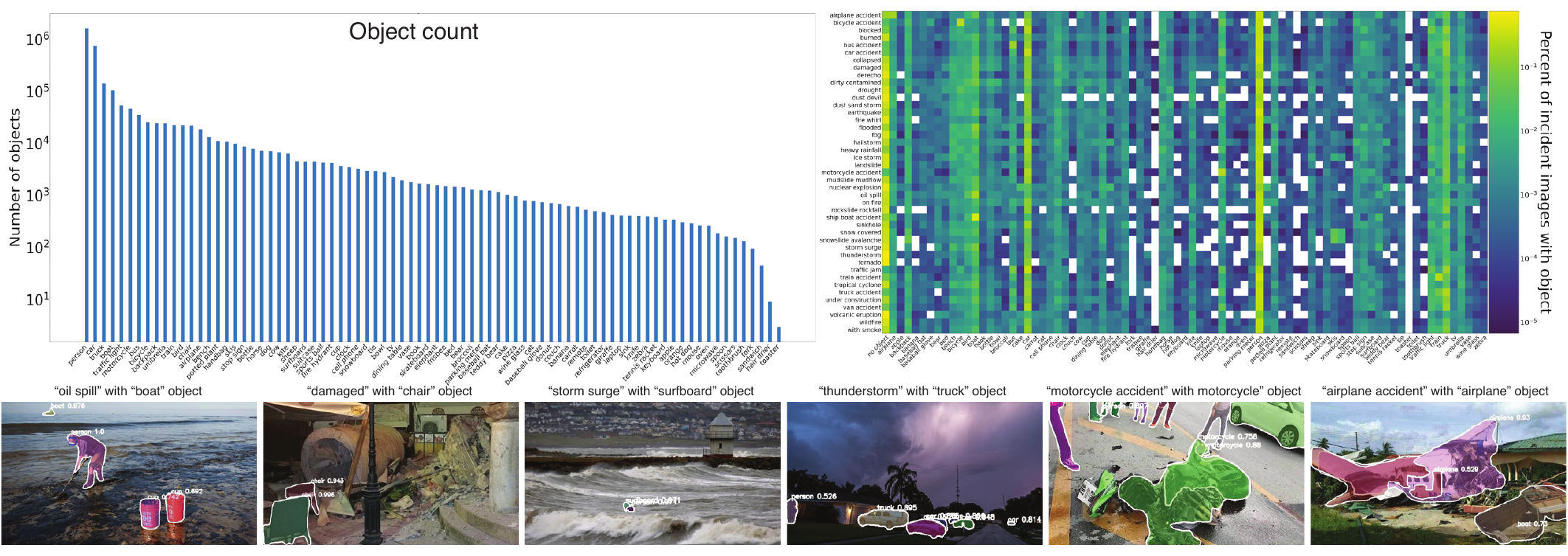}
\caption{\textbf{Objects in the Incidents1M Dataset.} We run a pretrained Mask R-CNN model on images with class-positive incidents. (Top left) We show the number of objects found in the entire dataset for a particular incident category. (Top right) We show a 2D histogram of incident images that contain at least one detected object. We normalize each row. (Bottom) We show qualitative examples of incidents with detected objects overlaid.}
\label{fig:object_statistics}
\end{figure*}

\begin{table}[t]
\begin{center}
\caption{\textbf{Class-negative loss and dataset size.} In this table, we explore the importance of the class-negative loss (top section) as well as the dataset size (middle section). The final row is the model used in the original Incidents work~\cite{weber2020detecting} (bottom section). Our best performing model is row 2, which uses both the class-positives and class-negatives; this is the Incidents1M Model chosen for in-the-wild experiments.}
\begin{tabular}{ccccc}
\toprule
\multicolumn{1}{c}{} & \multicolumn{2}{c}{\textbf{Loss}} & \multicolumn{2}{c}{} \\
\multicolumn{1}{c}{\textbf{Training data}} & \multicolumn{1}{c}{\textbf{Cls-pos}} & \multicolumn{1}{c}{\textbf{Cls-neg}} & \multicolumn{2}{c}{\textbf{Test set mAP}} \\ \midrule
\multicolumn{1}{c}{} &  &  & \multicolumn{1}{c}{Incident} & \multicolumn{1}{c}{Place} \\
\multicolumn{1}{l}{Incidents1M} & \multicolumn{1}{c}{\checkmark} &  & \multicolumn{1}{c}{62.83} & \multicolumn{1}{c}{63.15}  \\
\multicolumn{1}{l}{Incidents1M} &  \multicolumn{1}{c}{\checkmark} &  \multicolumn{1}{c}{\checkmark} & \multicolumn{1}{c}{\textbf{67.19}} & \multicolumn{1}{c}{62.94}  \\ \midrule
\multicolumn{1}{l}{Incidents1M (75\%)} &  \multicolumn{1}{c}{\checkmark} &  \multicolumn{1}{c}{\checkmark} & \multicolumn{1}{c}{66.85} & \multicolumn{1}{c}{62.52}  \\
\multicolumn{1}{l}{Incidents1M (50\%)} &  \multicolumn{1}{c}{\checkmark} &  \multicolumn{1}{c}{\checkmark} & \multicolumn{1}{c}{65.59} & \multicolumn{1}{c}{61.57}  \\
\multicolumn{1}{l}{Incidents1M (25\%)} &  \multicolumn{1}{c}{\checkmark} &  \multicolumn{1}{c}{\checkmark} & \multicolumn{1}{c}{64.01} & \multicolumn{1}{c}{59.68}  \\ \midrule
\multicolumn{1}{l}{Incidents~\cite{weber2020detecting}} & \multicolumn{1}{c}{\checkmark} & \multicolumn{1}{c}{\checkmark} & \multicolumn{1}{c}{63.89} & \multicolumn{1}{c}{60.33} \\ \bottomrule
\end{tabular}
\end{center}
\label{table:class_negatives_and_larger_dataset}
\end{table}

\subsection{More data improves performance}

We note that in the original Incidents Dataset \cite{weber2020detecting}, only single-labels were used with a similar loss as Eq.~\ref{eq:loss}. However, in this paper, we have both (1) multiple labels and (2) more images. Here we show the effect of scaling up the dataset in the middle rows of Tab.~\ref{table:class_negatives_and_larger_dataset}. The 2nd row of the table is the full dataset (100\%). The very bottom row is the trained model from \cite{weber2020detecting}, which has worse performance compared to our model trained on our expanded Incidents1M dataset. Note that the same model architecture and loss is used in both papers.

Below row 2 in Tab.~\ref{table:class_negatives_and_larger_dataset}, we show the impact that dataset size has on model performance. In rows 3-5, we show the results of models trained with 75\%, 50\%, and 25\% of the Incidents1M dataset. The subsets are randomly sampled from all the 1,787,154 images, and we ensure that each smaller set is a strict subset of the larger one. The mAP gracefully declines with an incident mAP from 67.19 (100\%) to 64.01 (25\%) and a place mAP from 62.94 (100\%) to 59.68 (25\%). We notice that even when using just 25\% (446,788 images), we have comparable performance to the original model~\cite{weber2020detecting} shown in the last row. That model was trained with 1,144,148 images where 446,684 contained exactly one class-positive image and the remaining 697,464 images only had class-negative information. From this, we observe that the multi-label information in the Incidents1M dataset is helpful to model performance.

The Incidents1M Model used in the remainder of this paper is from row 2 of Tab.~\ref{table:class_negatives_and_larger_dataset}, the best model for incident detection, trained with both class-positives and class-negatives on the entire dataset. Although place mAP is not optimal, it has comparable performance to the class-positive-only loss and nevertheless, throwing away useful class-negative information for training models is counter-productive to our detection goals.

% ########################################################
\section{Incidents1M Dataset bias analysis}

In this section, we analyze the Incidents1M Dataset using an off-the-shelf pretrained detection model to provide further insights into our data. This is similar to \cite{wang2020revise}, where we aim to identify sources of bias in the data rather than in only the model performance. We also examine where the images may be coming from by geo-locating IP addresses, which can be used as a proxy for location. Lastly, we run our Incidents1M Model on a set of images across continents to discover model performance biases based on geography.

\subsection{Detecting objects and understanding correlations}
We run a Mask R-CNN~\cite{he2017mask} model pretrained on 80 COCO categories over all of the class-positive incident images in the Incidents1M Dataset. We filter all object detections above a 0.75 confidence threshold, and we report our findings in Fig.~\ref{fig:object_statistics} (left). We observe that ``person" and ``car" are the most common object categories. Furthermore, we observe which objects occur with which incidents. We notice that indoor objects (e.g., ``couch" and ``microwave") tend to be false positive detections or correct predictions that appear out of place due to an incident such as an earthquake or flood, which may destroy a home and cause indoor objects to appear outside. The class-positive images in the Incidents1M Dataset have an average of 4.92 and a median of 3 objects.

On the top right of Fig.~\ref{fig:object_statistics}, we look for correlations of incidents occurring with different objects. The 2D histogram shows the percent of incident images that contain at least one instance of the specified object. The colors are row-wise normalized. Some obvious examples that occur are ``traffic jam" with ``car", ``ship boat accident" with ``boat", and ``bus accident" with ``bus". However, more interesting analysis is found by looking at a particular incident and looking for relative peaks in the corresponding row. For example, ``tornado" incidents occur with ``boat", ``car", ``kite", ``traffic light", ``truck", and ``umbrella". We also notice that most images, regardless of the incident, occur with ``no object", ``car", ``person", and ``truck". This may indicate that most images were taken on the road or in public areas, which may match expectations given that the original dataset images are downloaded from the Google search engine.

At the bottom of Fig.~\ref{fig:object_statistics}, we include qualitative examples of incidents occurring with specific Mask R-CNN detections indicated. By using an off-the-shelf model to detect objects, we can enable retrieving images of interest, e.g., all the images that contain a ``thunderstorm" incident with a ``truck" object. This could be helpful when training a class-specific model or for analyzing specific damage types, such as how ``car" objects are affected by ``flood" incidents.

\newcolumntype{?}{!{\vrule width 1pt}}
\begin{table*}[t]
\begin{center}
\caption{\textbf{Geographical model bias.} (Left) We show the mAP of the Incidents model across 6 different continents for 14 incident super-categories. Each country is represented with 1K geotagged images from Twitter. We also show the results on Incidents1M test set for reference. (Right) We show the number of class-positive images for each super-category and the total below (where an image has at least one incident).}
\begin{tabular}{lccccccc||lcccccc}
\toprule
 & \multicolumn{6}{c}{\textbf{Average Precision (AP)}} & & & \multicolumn{6}{c}{\textbf{Number of class-positive images}} \\
Super-category & {AF} & {AS} & {EU} & {NA} & {OC} & {SA} & {Incident1M} & Super-category & {AF} & {AS} & {EU}         & {NA}         & {OC}         & {SA}         \\ \midrule
blocked & 5.73 & 12.63 & 37.34 & 29.96 & 14.01 & 14.17 & 67.53         & blocked & 33 & 68 & 34 & 75 & 59 & 113 \\
burned- & 71.24 & 70.78 & 55.05 & 56.22 & 53.86 & 37.53 & 75.95         & burned-      & 163 & 118          & 110 & 119 & 108 & 37                             \\
collapsed- & 38.46 & 48.39 & 41.70 & 36.39 & 44.40 & 67.03 & 61.00  & collapsed-      & 166 & 164          & 95 & 121 & 154 & 325                            \\
dirty- & 35.38 & 28.69 & 24.00 & 23.01 & 29.57 & 32.09 & 64.81         & dirty-     & 103 & 128          & 29 & 75 & 64 & 118                            \\
drought & 66.37 & 31.16 & 39.31 & 29.74 & 47.21 & 41.05 & 70.51         & drought & 54 & 24 & 36 & 35 & 40 & 39                             \\
earthquake & 68.15 & 76.88 & 84.00 & 69.32 & 49.85 & 85.44 & 79.59         & earthquake  & 61 & 118          & 46 & 51 & 78 & 260                            \\
flooded & 87.47 & 83.29 & 76.55 & 72.26 & 75.93 & 55.65 & 86.76         & flooded & 145 & 168          & 109 & 85 & 106 & 97                             \\
fog & 42.11 & 61.47 & 60.68 & 50.66 & 38.10  & 40.90  & 79.54         & fog   & 43 & 76 & 92 & 103 & 158 & 62                             \\
rainfall & 54.25 & 62.75 & 44.58 & 45.54 & 50.02 & 29.81 & 65.95         & rainfall         & 147 & 241          & 116 & 101 & 163 & 78                             \\
landslide & 13.84 & 37.22 & 15.87 & 17.98 & 26.95 & 41.71 & 58.85         & landslide   & 17 & 63 & 20 & 65 & 34 & 127                            \\
on fire & 86.43 & 73.95 & 70.85 & 78.91 & 58.22 & 65.82 & 69.23         & on fire & 133 & 82 & 82 & 79 & 63 & 17                             \\
snowstorm- & 50.53 & 49.66 & 70.30  & 58.92 & 22.64 & 51.81 & 55.30          & snowstorm- & 11 & 16 & 63 & 64 & 27 & 16                             \\
traffic jam & 50.20  & 47.67 & 76.69 & 64.25 & 55.31 & 65.50  & 75.42         & traffic jam & 22 & 22 & 8  & 12 & 8  & 19                             \\
veh. accident & 57.26 & 52.44 & 51.46 & 46.51 & 45.58 & 31.63 & 68.77         & veh. accident       & 84 & 42 & 31 & 43 & 37 & 73                             \\ \midrule
mAP & 51.96 & 52.64 & 53.46 & 48.55 & 43.69 & 47.15 & 69.94         & Total of 1K          & 580 & 629 & 534 & 535 & 594 & 607 \\ \bottomrule
\end{tabular}
\end{center}
\label{table:geo_model_bias}
\end{table*}

\subsection{Geographical distribution}
\begin{figure}[t]
\centering
\includegraphics[width=\linewidth]{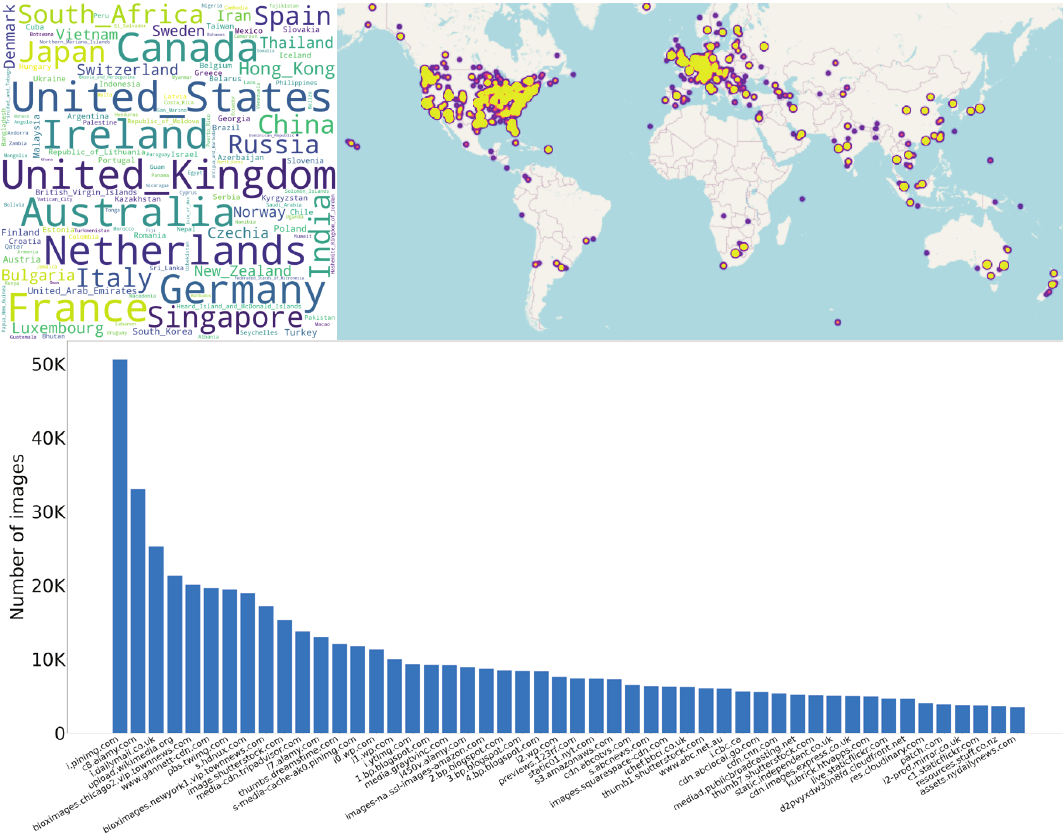}
\caption{\textbf{Geographical distribution.} Here we report the approximate locations of images according to URLs. We perform geolocation with queries to \url{https://geolocation-db.com}. (Top) A word cloud and heatmap indicating where most of the images are located. (Bottom) A histogram of the 50 most common domain names with the corresponding number of images from the geolocalized server.}
\label{fig:geo_dist}
\end{figure}

In this section, we conduct a brief geographical analysis by approximating the latitude and longitude coordinates of each image based on its URL used to download the image. We note that this is an approximation of the location coordinates for two reasons: (1) we are approximating server locations rather than where the images were actually taken and (2) the database mapping server IP addresses to GPS coordinates may be outdated or incorrect. For each image URL, we first extract the domain name and then use DNS to map it to an IP address. Then, we query \url{https://geolocation-db.com} with the IP address to receive an approximated location. In Fig.~\ref{fig:geo_dist}, we report visualizations for where most of the images are located\footnote{World maps in the figures are plotted with Plotly and are enabled by \href{https://www.openstreetmap.org}{OpenStreetMap}.}. At the top, we see that most images are stored on servers in the United States and Europe. At the bottom in the histogram, we notice that the image locations follow a long-tail distribution. There are only 177,656 unique domains out of the 1.8M URLs (\textapprox10\%). In the histogram, we report the 50 most common domains.

\subsection{Geographical model bias experiment}

Here we aim to investigate if our model exhibits bias based on geographical location of where images are taken. To do this, we use Twitter; we download images associated with geotagged tweets that have natural disaster keywords. From this collection, we separate the tweets into 6 continents. We exclude Antarctica (AN) from our analysis as tweets are not common there. Once separated into continents with tweet GPS metadata, we then obtain ground-truth labels from MTurk. To exhaustively evaluate the models, we would have to obtain labels for all 43 incidents, which quickly becomes costly.

To speed up analysis, we first group our incident categories into the following 14 super-categories: ``blocked", ``burned-with smoke", ``collapsed-damaged", ``dirty-contaminated", ``drought", ``earthquake", ``flooded", ``fog", ``heavy rainfall", ``landslide", ``on fire", ``snowstorm-snow covered", ``traffic jam", and ``vehicle accident". Next, we sample 1K images from each continent and for each image, we ask 14 binary questions (i.e., Does this image contain X incident?). Finally, we use these labels to obtain a mean average precision (mAP) metric shown in the left of Tab.~\ref{table:geo_model_bias}. We include the AP on the Incidents1M Dataset as well for reference. On the right of the table, we show the number of class-positive incident images for the particular super-categories as well as the total number of class-positive images out of the 1K images per continent. Recall that a class-positive image has at least one incident.

\subsubsection{Average precision (AP)}

We notice that the mAP for each continent is much lower than the mAP on the Incidents1M Dataset (69.94). Europe (EU) has the highest mAP of 53.46, which makes sense given our findings in Fig.~\ref{fig:geo_dist} where images from Incidents1M are often located on servers in EU and North America (NA). However, this logic does not apply for NA, where the mAP is only 48.55. We suspect that although images are often located in NA, they are often of incidents located elsewhere. Maybe geolocalization work such as \cite{lin2013cross} could prove useful to study where incidents are occurring based on their visual features rather than using Twitter-provided GPS coordinates.

\begin{figure*}[t]
\centering
\includegraphics[width=\linewidth]{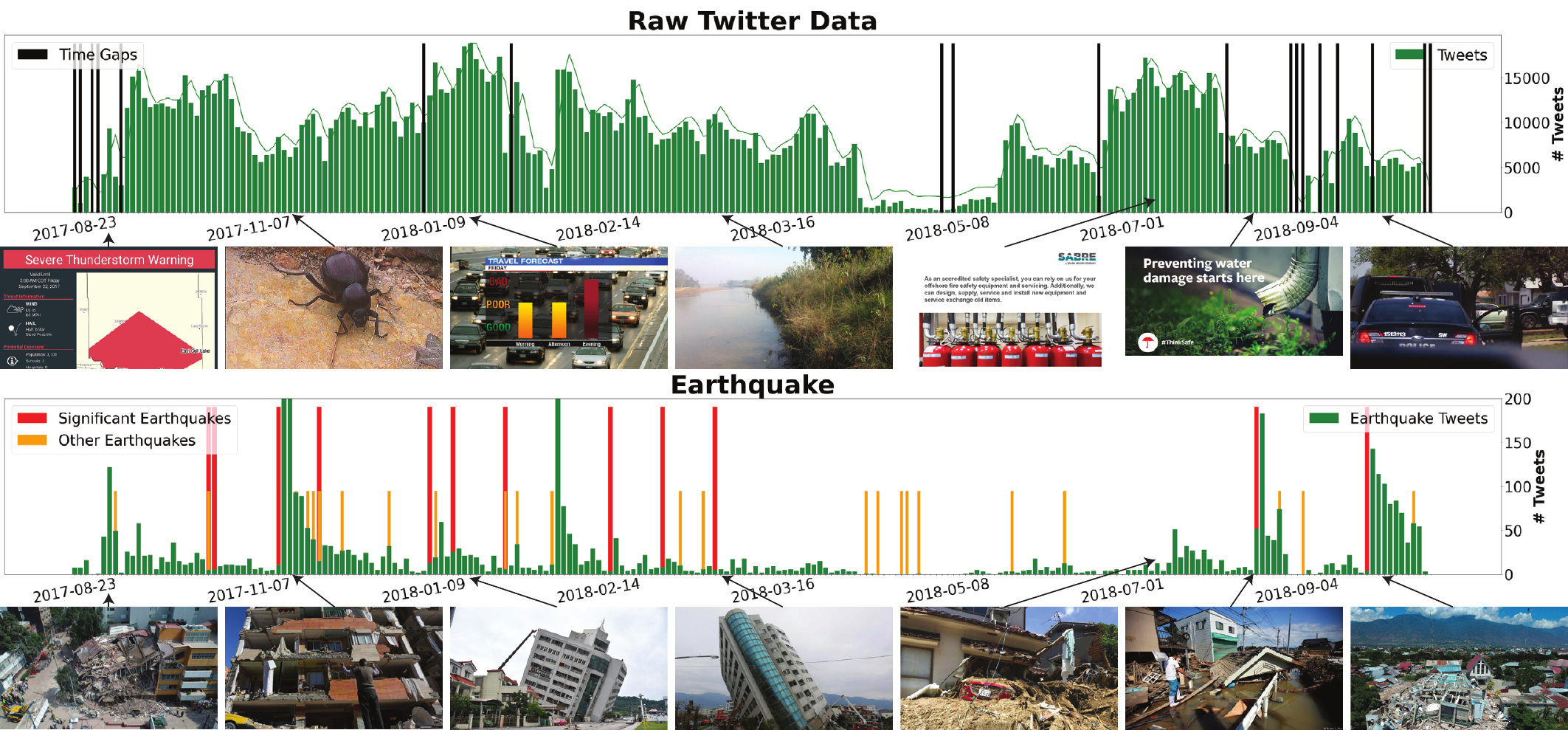}
\caption{\textbf{Temporal monitoring on Twitter.} (Top) Histogram of tweets obtained from Twitter using natural disaster keywords from 2017-2018. Black lines indicate periods of time when our data collection server was inactive. (Bottom) Number of tweets with earthquake images per day after filtering with at least 0.5 confidence. For significant earthquakes (above 6.5 magnitude), we notice an increase in earthquake images immediately after the event. Furthermore, we notice a spike on July 20, 2018 not reported in the NOAA database. We manually checked the tweets and found images referring to a severe flood in Japan, indicating that the flood damage may resemble earthquake damage.}
\label{fig:twitter_temporal_earthquake}
\end{figure*}

We see that even the best mAP from EU (53.46) is still 16.48 mAP points below the score on the Incidents1M Dataset. This is due to Twitter images being out-of-distribution and having more challenging negative images to filter through than our Incidents1M test set. We include the per super-category APs in the table for additional inspection. We notice that for some categories such as ``earthquake", ``on fire", and ``snowstorm", the AP scores are comparable for the continents (except for OC) and with Incidents1M. For others such as ``blocked", ``dirty-contaminated", and ``landslide", the APs are low for all continents compared to Incidents1M.

\begin{figure}[t]
\centering
\includegraphics[width=\linewidth]{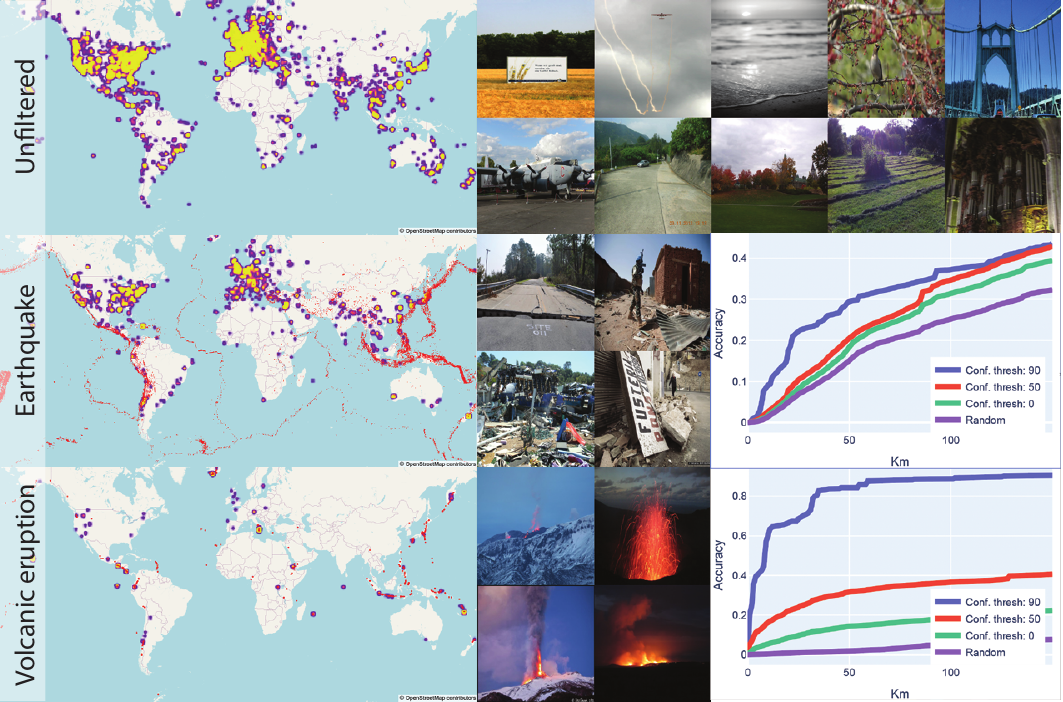}
\caption{\textbf{Incident detection on Flickr images.} Here we show incident detection results on 26M Flickr images. (Top) On the left we show a heatmap of where the images are located, and on the right we show some images. Red dots are earthquake epicenters or volcano locations. (Middle) we show locations of images filtered with high confidence for earthquakes. These correspond to the $0.9$ threshold (blue lines) in the graphs. We filter at different thresholds and show Accuracy@XKm (percent of images within X kilometers from a red dot) increasing. (Bottom) The same but for volcanic eruptions.}
\label{fig:flickr_filtering}
\end{figure}

\subsubsection{Distribution of class-positive images}

On the right of Tab.~\ref{table:geo_model_bias}, we see the breakdown of per super-category class-positive images followed by the total at the bottom. First, we notice that some super-categories occur quite often in Twitter images (e.g., ``burned-with smoke", ``collapsed-damaged", ``rainfall") while others occur less frequently (e.g., ``snowstorm-snow covered", ``traffic jam"). We also notice local peaks in frequency depending on the continents. Unexpectedly, the EU and NA have the most ``snowstorm-snow covered" labels. Africa (AF) has the most ``burned-with smoke" and ``on fire" labels. AF and Asia (AS) have the most ``flooded" images. We see that AS has the most class-positive incident images (629) out of the 1K images labeled. We note that the categories such as ``vehicle accident" with low counts could be due to the way the original Twitter images were downloaded (i.e., based on natural disaster keywords).

% ########################################################
\section{Incident detection in the wild}
\label{sec:detection_applications}

In this section, we use our Incidents1M Dataset model and repeat experiments from \cite{weber2020detecting} for incident detection in both Flickr and Twitter images. We perform detection experiments on Flickr and Twitter. We also show a temporal monitoring experiment where we filter Twitter data over time to detect incidents, and we comment on the usefulness of such a pipeline with a custom interface built internally. We note that in all filtering experiments that we repeat here, our Incidents1M Model has performance metrics greater or equal to that of our prior work \cite{weber2020detecting}.
%
% \dimpp{general comment about 7.1-7.3. I would also include the results on the ECCV paper for comparison and add just a sentence to compare the results in each subsection. Similar to Tab 1 this will show that the new model performs better in all these experiments. eg. Tab 3 and Tab in Fig 9 can show the filtered numbers from ECCV. It's hard to add these in Fig 7,8, unless you can a good idea how to do it easily}

\begin{figure*}[t]
\centering
\includegraphics[width=\linewidth]{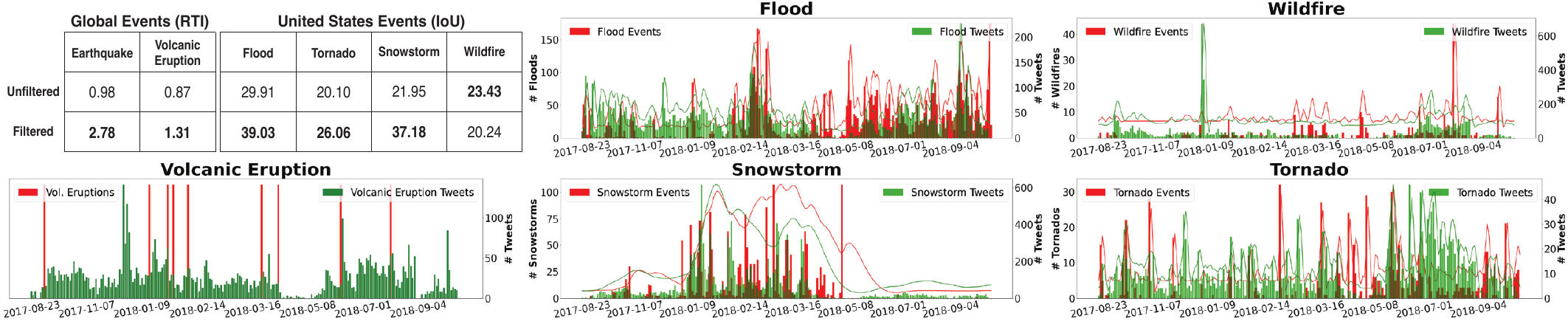}
\caption{\textbf{Temporal monitoring on Twitter.} (Top left) We report mRTI for global events and IoU for common US events. (Bottom left) We show a similar result as in Fig.~\ref{fig:twitter_temporal_earthquake} but for filtered volcanic eruptions images and ground truth events. (Right) For more frequent events in the United States, we filter tweets for flood, tornado, snowstorm, and wildfire images and compare with ground truth frequency events obtained from NOAA.}
\label{fig:twitter_temporal_most}
\end{figure*}

\subsection{Incident detection on Flickr}
The goal of this experiment is to illustrate how our model can be used to detect specific incident categories in the wild. Here we use 26 million geo-tagged Flickr images obtained from the YFCC100M dataset~\cite{thomee2016yfcc100m}. Since the images have precise geo-coordinates from EXIF data, we can use our incident detection model to filter for specific incidents and compare distance to ground-truth locations. We evaluate only earthquake and volcanic eruption incidents in this experiment as we could find reasonable ground-truth data to compare the results. Specifically, we downloaded the GPS coordinates, i.e., latitude and longitude, of a public compilation of earthquake epicenters\footnote{\url{https://raw.githubusercontent.com/plotly/datasets/master/earthquakes-23k.csv}} and volcanoes from the National Oceanic and Atmospheric Administration (NOAA) website\footnote{\url{https://www.noaa.gov/}}. We employ an Accuracy@XKm metric~\cite{accuracy_x_km} to determine whether the predicted incident is correct or not. More concretely, we compute the percentage of images within X Km from the closest earthquake epicenter or volcano, respectively. We randomly sample images and report metrics for (i) unfiltered images, (ii) images with the target incident having the highest score, (iii) images with model confidence above $0.5$, and (iv) images with model confidence above $0.9$. Fig.~\ref{fig:flickr_filtering} shows that detected earthquake and volcanic eruption incidents appear much closer to expected locations when compared to random images.

\subsection{Incident detection on Twitter}

In this experiment, we detect earthquakes and floods in noisy Twitter data posted during actual disaster events. We collected data from four earthquake and two flood events using event-specific hashtags and keywords. In total, 901,127 images were downloaded. Twitter GPS coordinates are not nearly as precise as the Flickr ones, so we consider only the 38,317 geo-located images within 250 Km from either (i) the earthquake epicenter or (ii) the flooded city center.

For all six events shown in Tab.~\ref{table:twitter_detection}, we use MTurk to obtain ground-truth human labels (i.e., earthquake or not, and flood or not) for images within the considered radius. Then, we compare the quality of the initial set of the keyword-based retrieved Twitter images (unfiltered) to the quality of images retained by our model (filtered). We report the average precision (AP) per event for both earthquakes and floods. When considering all earthquake events and flood events, we obtain an average AP of 77.71\% and 92.96\% compared to the baseline AP of 15.87\% and 35.68\%, respectively. The baseline AP is the AP averaged over multiple trials of randomly shuffling the images, and it is given as a reference.

\begin{table}[t]
\begin{center}
\caption{\textbf{Twitter incident detection.} Here we demonstrate incident detection on Twitter images by sorting with confidence scores from our model and reporting average precision (filtered AP). We compare against a randomly sorted baseline (unfiltered AP). We consider all tweets within a 250 Km radius of the earthquake epicenters (for earthquakes) or from the center of the city (for floods). Ground-truth labels are obtained from MTurk for each event to compute the AP for both the unfiltered and filtered scenarios.}
\begin{tabular}{@{}ccccc@{}}
\toprule
\multicolumn{1}{@{}l}{\textbf{Incident}} & \multicolumn{1}{c}{\textbf{Location}} & \multicolumn{1}{c}{\textbf{Number}} & \multicolumn{2}{c@{}}{\textbf{AP}} \\
\multicolumn{1}{c}{} & \multicolumn{1}{c}{\textbf{\& Year}} & \multicolumn{1}{c}{\textbf{of images}} & \multicolumn{1}{c}{\textbf{Unfiltered}} & \multicolumn{1}{c@{}}{\textbf{Filtered}} \\ \midrule
\multicolumn{1}{@{}l}{Earthquake} & \multicolumn{1}{c}{Nepal 2015} & \multicolumn{1}{c}{2479} & \multicolumn{1}{c}{22.30} & \multicolumn{1}{c@{}}{83.57} \\
\multicolumn{1}{@{}l}{Earthquake} & \multicolumn{1}{c}{Chile 2015} & \multicolumn{1}{c}{2015} & \multicolumn{1}{c}{8.92} & \multicolumn{1}{c@{}}{62.97} \\
\multicolumn{1}{@{}l}{Earthquake} & \multicolumn{1}{c}{Ecuador 2016} & \multicolumn{1}{c}{5659} & \multicolumn{1}{c}{25.37} & \multicolumn{1}{c@{}}{89.32} \\
\multicolumn{1}{@{}l}{Earthquake} & \multicolumn{1}{c}{Italy 2016} & \multicolumn{1}{c}{18673} & \multicolumn{1}{c}{6.90} & \multicolumn{1}{c@{}}{74.98} \\ \midrule
\multicolumn{1}{@{}l}{Flood} & \multicolumn{1}{c}{Chennai 2015} & \multicolumn{1}{c}{5091} & \multicolumn{1}{c}{26.30} & \multicolumn{1}{c@{}}{93.25} \\
\multicolumn{1}{@{}l}{Flood} & \multicolumn{1}{c}{Bangladesh 2017} & \multicolumn{1}{c}{297} & \multicolumn{1}{c}{45.06} & \multicolumn{1}{c@{}}{92.66} \\ \bottomrule
\end{tabular}
\end{center}
\label{table:twitter_detection}
\end{table}

\subsection{Temporal monitoring of incidents on Twitter}
\label{sec:temporal_monitorning}

In this section we demonstrate how our model can be used on a Twitter data stream to detect specific incidents over time. To test this, we downloaded 1,946,850 images from tweets containing natural disaster keywords (e.g., blizzard, tornado, hurricane, earthquake, active volcano, coastal flood, wildfire, landslide) from Aug. 23, 2017 to Oct. 15, 2018. To quantify detection results, we obtained ground-truth event records from the ``Significant Earthquake Database", the ``Significant Volcanic Eruption Database", and the ``Storm Events Database" of NOAA. The earthquake and volcanic eruptions ground-truth events are rare \textit{global} events, while the storms (floods, tornadoes, snowstorms and wildfires) are much more frequent but recorded only for the \textit{United States}. We filter images with at least $0.5$ confidence and compare against the databases (Fig.~\ref{fig:twitter_temporal_earthquake} and Fig.~\ref{fig:twitter_temporal_most}).

For earthquakes and volcanic eruptions, we report average Relative Tweet Increase (RTI) inspired by~\cite{avvenuti2014ears}, where
\begin{equation}
    \textrm{RTI}_{e} = \frac{\sum_{d=e}^{e+w} N_{d}}{\sum_{d=e}^{e-w} N_{d}}
\end{equation}
$N_{d}$ is the number of relevant images posted on day $d$, $e$ is the event day (e.g., day of earthquake or volcanic eruption), and $w$ is an interval of days. We use $w=7$ for our analysis to represent a week before and after an event. An $RTI$ of 2 means that the average number of tweets in the week following an event is twice as high as the average number the week before. After filtering, the mean RTI ($\textrm{mRTI} = \sum_{e \in E} RTI_{e}/|E|$) shows an average of $2.78$ folds increase in tweets the week after an earthquake and $1.31$ folds after a volcanic eruption. The numbers are reported in Fig.~\ref{fig:twitter_temporal_most}). A more qualitative depiction for earthquakes is shown in Fig.~\ref{fig:twitter_temporal_earthquake}.

We notice that the mRTI would be even better if the ground truth databases were exhaustive; unfortunately, this is not the case. On Nov. 27, 2017 we detect the highest number of volcanic eruption images, but observe no significant eruption in the database. Looking into this, we found that Mount Agung erupted the same day, which caused the airport in Bali, Indonesia to close and left many tourists stranded\footnote{\url{https://en.wikipedia.org/wiki/Mount_Agung}}.

\begin{figure}[t]
\centering
\includegraphics[width=\linewidth]{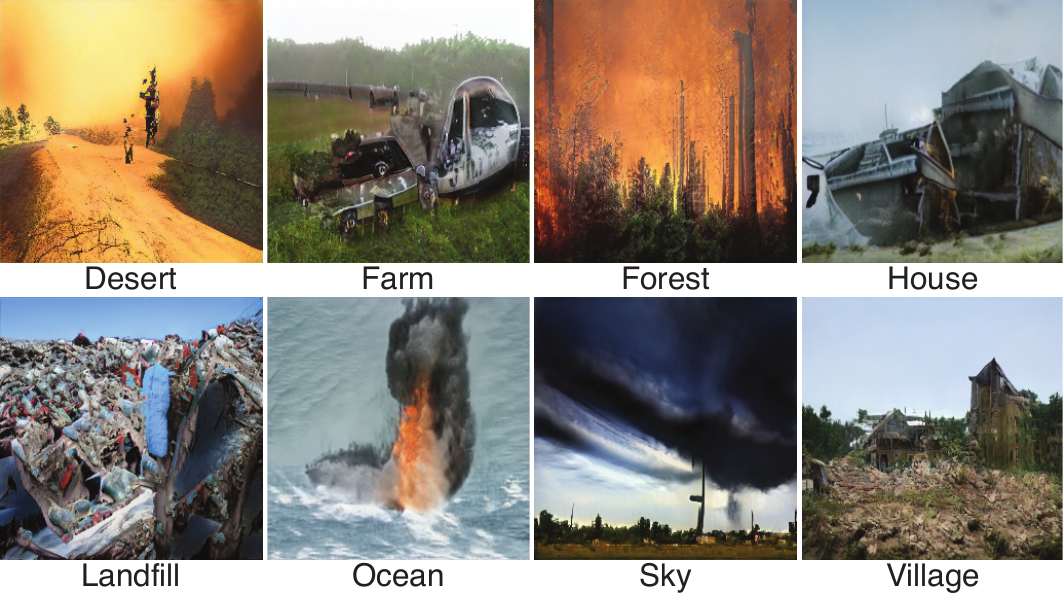}
\caption{\textbf{StyleGAN2-generated images.} Here are images generated from a places class-conditional StyleGAN2 model. The model is trained with the 764,124 images that have at least one class-positive place label. We do not condition on incident labels, yet random seed sampling in the latent space yields diverse, realistic looking images.}
\label{fig:stylegan}
\end{figure}

For the more common events (floods, wildfires, snowstorms, and wildfires), we measure the correlation between tweet frequency and event frequency. We normalize both histograms, smooth with a low-pass filter, and report intersection over union (IoU) for United States incidents in Fig.~\ref{fig:twitter_temporal_most}. We notice an increase in IoU after filtering for flood, tornado, and snowstorm images. For wildfires, we notice a decrease in IoU and attribute this to the large spike in tweets in December 2017. In fact, frequency of events does not necessarily capture how severe the damage is (which, in turn, is often correlated with how many people post on social media). In fact, a destructive wildfire occurred in California on Dec. 4, 2017 burning 281,893 acres\footnote{\url{https://en.wikipedia.org/wiki/Thomas_Fire}}.

\begin{figure}[t]
\centering
\includegraphics[width=\linewidth]{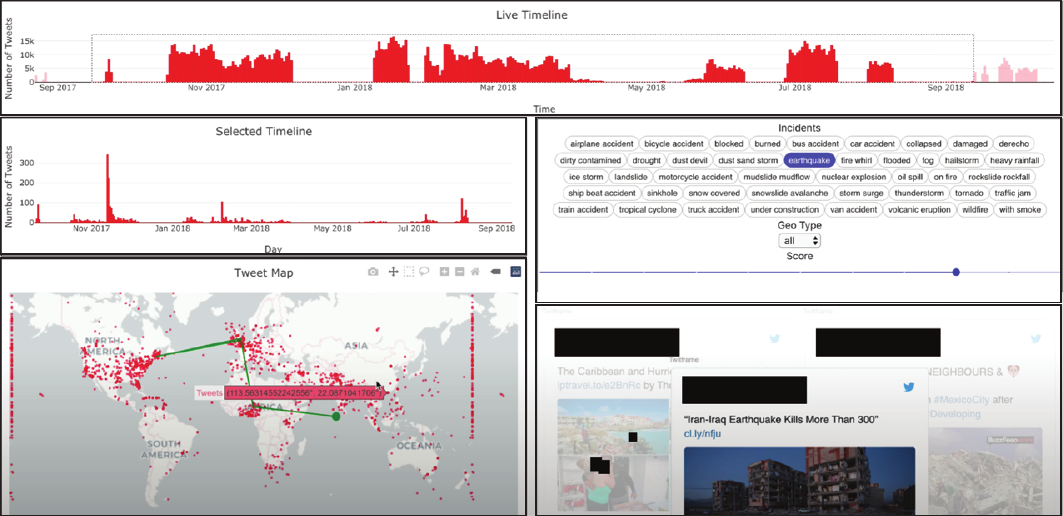}
\caption{\textbf{Interactive dashboard to monitor Twitter data.} We built an internal web-based JavaScript application to better understand the incidents that occur in images on Twitter. We show histograms (top and middle left) for the selected incident and confidence score filter (middle right). (Bottom left) we show where these tweets are located on a world map and (bottom right) we show the actual tweets as HTML IFrames. Connected lines in the map indicate which tweets and where re-tweets occurred, with the green dot indicating the original tweet.}
\label{fig:interactive_dashboard}
\end{figure}

In this experiment we show results in an off-line setting with a collection of 1,946,850 tweets from the past. Our straightforward procedures could be implemented to work in real-time with temporal predictions or emergency alerts when high mRTI peaks are detected. To better understand this idea and improve interpretation of this large tweet collection, we have built an internal interface shown in Fig.~\ref{fig:interactive_dashboard} and discuss it in Sec.~\ref{sec:interactive_dashboard}. Our goal is to demonstrate that the Incidents1M Model can offer real-world insights when operating on large, unstructured social media platforms such as Twitter.

% ########################################################
\section{Additional applications of Incidents1M}
\label{sec:additional_applications}
In this section, we aim to go beyond the incident detection paradigm in the previous sections and instead explore some additional applications of the Incidents1M Dataset. In particular, we (1) train a StyleGAN2 model~\cite{Karras2019stylegan2, Karras2020ada} for incident image generation and (2) implement a JavaScript-based web application for understanding incidents that can be detected in Twitter data. We hope both of these ideas and applications inspire more work in the incident-analysis direction.

\subsection{Incident image generation}
Image generation methods have been popularized with advances in GAN models~\cite{goodfellow2014generative}, which can now synthesize very realistic images. Here we use the Incidents1M Dataset to train a StyleGAN2 model with adaptive discriminator augmentation (ADA)~\cite{Karras2020ada} with image generation results shown in Fig.~\ref{fig:stylegan}. More specifically, we train a class-conditional model that is conditioned on the place label. 764,124 images in Incidents1M have at least one class-positive place, so during training we sample images from this subset and randomly choose one of the place labels to condition with. We found that class-conditioning helps to produce realistic images that have a coherent-looking structure (e.g., fields are planar, water has a wavy texture, and buildings are mostly intact). The incident labels are not used for conditioning, yet randomly sampling from the latent space gives interesting and diverse images. We believe further work in this direction could improve understanding of the composition of incidents and how damage may play a role in destroying and deforming person-made structures such as houses or villages, like those shown in Fig.~\ref{fig:stylegan}.

\subsection{Interactive dashboard for incident monitoring}
\label{sec:interactive_dashboard}
We built an internal, interactive web-based dashboard for understanding incidents that occur in Twitter data. Fig.~\ref{fig:interactive_dashboard} shows a screenshot of our application which was implemented as a ReactJS\footnote{\url{https://reactjs.org}} application and uses the the Plotly JavaScript Open Source Graphing Library\footnote{\url{https://plotly.com/javascript}}. For the image shown in the figure, we are using the same data collection from Sec.~\ref{sec:temporal_monitorning}. We filter for earthquake images in this particular example, and we provide a detailed caption of the layout with Fig.~\ref{fig:interactive_dashboard}. We found this interactive web-interface very helpful for our understanding of incidents that occur in-the-wild, and we hope this initial work leads to more ideas or an open-sourced implementation to explore data in real-time and with data streams other than from Twitter.

% ########################################################
\section{Conclusion}\label{sec:conclusions}

In this paper, we extend the original Incidents dataset~\cite{weber2020detecting} to a larger and multi-label dataset. We analyze the composition and possible biases that exist in the data, perform in-the-wild incident detection experiments on social media with Flickr and Twitter images, and present preliminary experiments that go behind model performance and incident detection. The Incidents1M Dataset has 977,088 images with at least one incident and an additional 810,066 images that have useful class-negative information. This data can be used to train models for incident classification, class-conditional generation models such as StyleGAN2, or more. We expect our data release, analysis in model performance, and applications in-the-wild will enable and inspire more work for humanitarian aid. Computer vision could be used as an important tool for scientists as we navigate the coming years with more frequent natural disasters and attempt to mitigate the harmful effects of climate change.

% ########################################################
\section*{Acknowledgments}
This work was supported by the CSAIL-QCRI collaboration project and the RTI2018-095232-B-C22 grant from the Spanish Ministry of Science, Innovation and Universities. We thank Nuria Marzo and Aritro Biswas for their contributions on starting and pioneering the initial directions of the Incident project. We also thank Thomas Chen for participating in discussions and for preliminary analysis of biases and correlations in the dataset.



\ifCLASSOPTIONcaptionsoff
  \newpage
\fi

% uncomment this to inspect all citations!
% \nocite{*}

\bibliographystyle{IEEEtran}
\bibliography{bib.bib}


\begin{IEEEbiography}[{\includegraphics[width=1in,height=1.25in,clip,keepaspectratio]{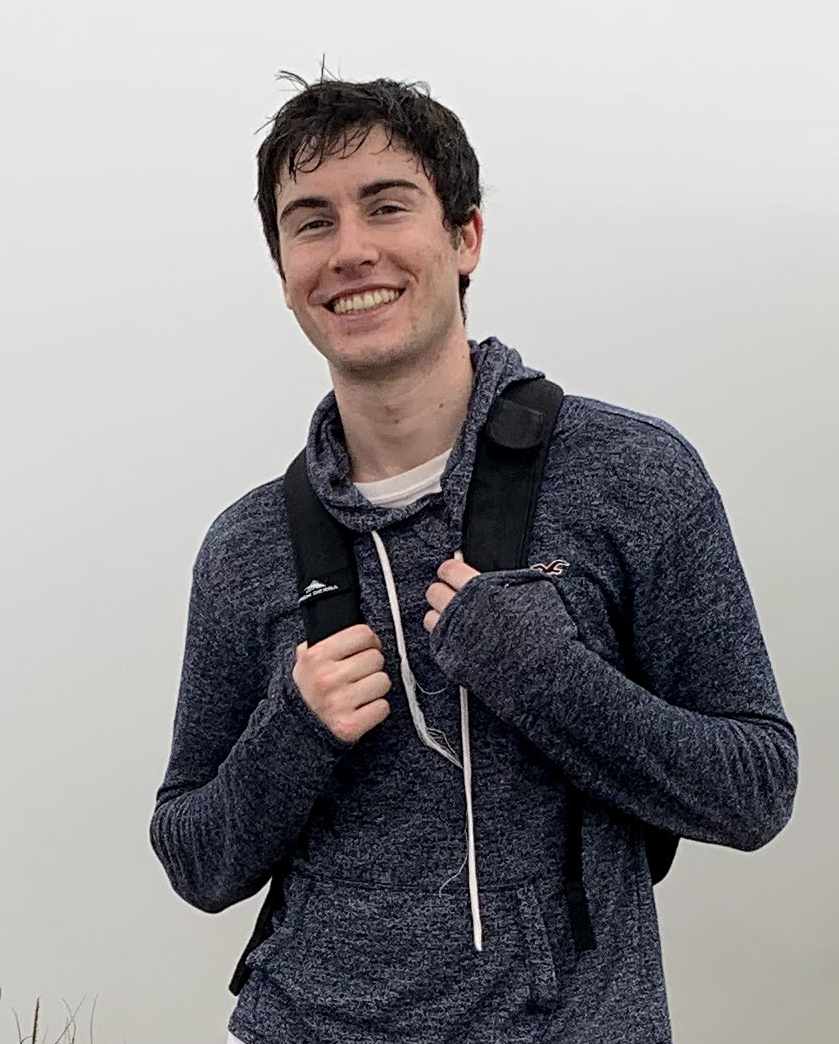}}]{Ethan Weber} completed his BS and MEng in EECS at MIT in '20 and '21, respectively. He completed the MEng under the guidance of Professor Antonio Torralba. Ethan has conducted and is interested in computer vision research with applications in robotics, augmented reality, 3D understanding, and humanitarian aid. He's now pursuing a PhD at The University of California, Berkeley at BAIR (Berkeley Artificial Intelligence Research Lab) where he's advised by Professor Angjoo Kanazawa. He documents his projects at \url{https://ethanweber.me}.
\end{IEEEbiography}

\begin{IEEEbiography}[{\includegraphics[width=1in,height=1.25in,clip,keepaspectratio]{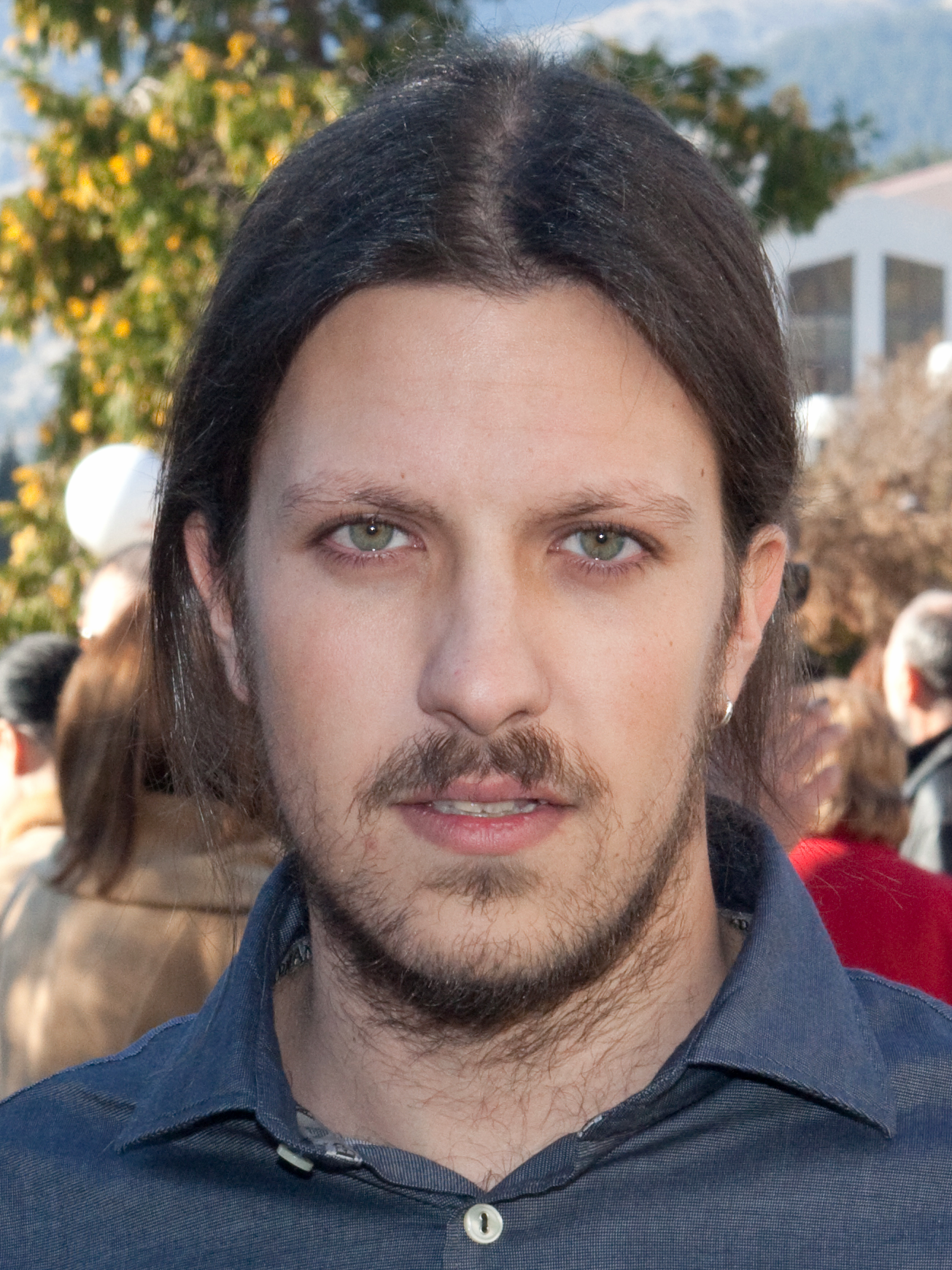}}]{Dim P. Papadopoulos} is a tenure-track Assistant Professor at the Department of Applied Mathematics and Computer Science at the Technical University of Denmark (DTU) since 2021. He received his MEng and MSc degrees in Electrical and Computer Engineering from Democritus University of Thrace (DUTH), Greece in 2011 and 2013, respectively. He received his PhD degree in Computer Vision from the University of Edinburgh, UK in 2018. From 2018 to 2021, he was a Postdoctoral Associate at CSAIL, MIT, MA, USA. His research interests are related to computer vision and machine learning. He won the CVPR Outstanding Reviewer Award in 2021 and the best paper award at the EUREKA conference in 2010.
\end{IEEEbiography}

\begin{IEEEbiography}[{\includegraphics[width=1in,height=1.25in,clip,keepaspectratio]{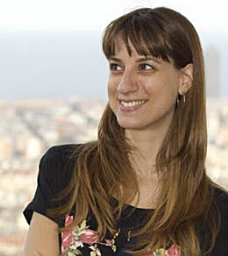}}]{Agata Lapedriza} is a Professor at Universitat Oberta de Catalunya (UOC) and a Research Affiliate at MIT Medialab. At UOC, she leads the Computer Vision group at UOC eHealth Research Center. Her research interests are related to Computer Vision, Natural Language Processing, Affective Computing, Explainable Artificial Intelligence (AI), and Fairness in AI. She is also interested in the applications of these research fields to Health, Social Robotics and, more generally, Artificial Intelligence for social good. She has been collaborating with MIT since 2012. From 2012 to 2015, she was a visiting professor at MIT CSAIL, where she worked on Scene Recognition and Interpretable Models in Computer Vision with Prof. Antonio Torralba. From 2017 to 2020, she was a visiting professor at MIT Medialab Affective Computing group, where she worked on emotion perception and social robotics with Prof. Rosalind Picard. More recently (2020-2021), she has been a visiting researcher at Google (USA). She did her PhD in Computer Science at the Universitat Autonoma de Barcelona and her BS degree in Mathematics at the Universitat de Barcelona.
\end{IEEEbiography}

\begin{IEEEbiography}[{\includegraphics[width=1in,height=1.25in,clip,keepaspectratio]{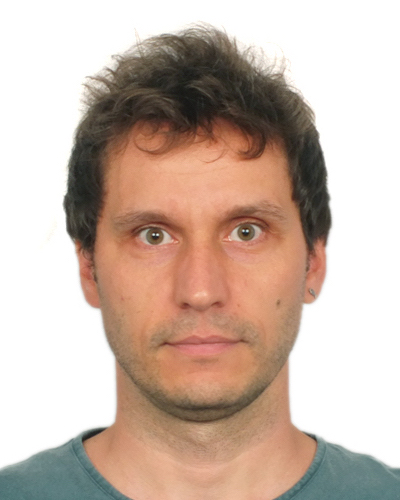}}]{Ferda Ofli} is a Senior Scientist at the Qatar Computing Research Institute since 2014. He received the B.Sc. degrees both in electrical and electronics engineering and computer engineering, and the Ph.D. degree in electrical engineering from Koc University, Istanbul, Turkey, in 2005 and 2010, respectively. From 2010 to 2014, he was a Postdoctoral Researcher at the University of California, Berkeley, CA, USA. His research interests cover computer vision and machine learning with applications in the humanitarian domain. He is an IEEE and ACM senior member with over 60 publications in refereed conferences and journals including CVPR, ECCV, and PAMI. He won the CVPR Outstanding Reviewer Awards in 2020 and 2021, the ISCRAM Best CoRe Paper Award in 2020, the ISCRAM Best Insight Paper and Best Paper Runner-up Awards in 2019, the ISCRAM Best Paper Runner-up Award in 2017, the Elsevier JVCI Best Paper Award in 2015, and the IEEE SIU Best Student Paper Award in 2011. He also received the Graduate Studies Excellence Award in 2010 for his outstanding academic achievement at Koc University.
\end{IEEEbiography}

\begin{IEEEbiography}[{\includegraphics[width=1in,height=1.25in,clip,keepaspectratio]{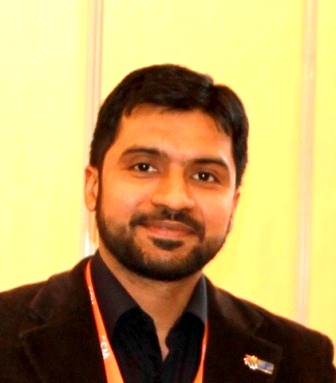}}]{Muhammad Imran}
Muhammad Imran is a Senior Scientist and Lead of the Crisis Computing team at Qatar Computing Research Institute (QCRI). His interdisciplinary research focuses on natural language processing, social computing, applied machine learning, human-computer interaction, and stream processing areas. He analyzes social media communications during time-critical situations using big data analysis techniques such as data mining, machine learning, and deep neural networks. He develops novel computational models, techniques, and technologies useful for stakeholders to gain situational awareness and actionable information during sudden onset disasters. Dr. Imran received his PhD in computer science from the University of Trento in 2013. He then joined QCRI as a post-doctoral researcher. Dr. Imran has published over 90 research papers in top-tier international conferences and journals including ACL, SIGIR, IJHCI, ICWSM, and WWW. Four of his papers received the "Best Paper Award" and two "Best Paper Runner-up Award." He has been serving as a co-chair of the Social Media Studies track of the ISCRAM international conference since 2014 and has served as Program Committee for many major conferences and workshops. Dr. Imran holds a Master of Science in Computer Science degree with a distinction from the Capital University of Science and Technology, Islamabad (2007).
\end{IEEEbiography}

\begin{IEEEbiography}[{\includegraphics[width=1in,height=1.25in,clip,keepaspectratio]{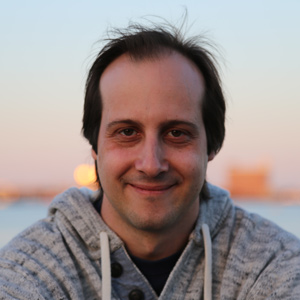}}]{Antonio Torralba} received the degree in telecommunications engineering from Telecom BCN, Spain, in 1994 and the Ph.D. degree in signal, image, and speech processing from the Institut National Polytechnique de Grenoble, France, in 2000. From 2000 to 2005, he spent postdoctoral training at the Brain and Cognitive Science Department and the Computer Science and Artificial Intelligence Laboratory, MIT. He is now a Professor of Electrical Engineering and Computer Science at the Massachusetts Institute of Technology (MIT). Prof. Torralba is an Associate Editor of the International Journal in Computer Vision, and has served as program chair for the Computer Vision and Pattern Recognition conference in 2015. He received the 2008 National Science Foundation (NSF) Career award, the best student paper award at the IEEE Conference on Computer Vision and Pattern Recognition (CVPR) in 2009, and the 2010 J. K. Aggarwal Prize from the International Association for Pattern Recognition (IAPR).
\end{IEEEbiography}

\end{document}